%% file: acl_latex.tex
\pdfoutput=1
\documentclass[11pt]{article}

\usepackage[]{acl}

\usepackage{times}
\usepackage{latexsym}
\usepackage{multirow}
\usepackage{verbatim}
\usepackage{graphicx}
\usepackage{makecell}
\usepackage{amssymb}

\usepackage[T1]{fontenc}
\usepackage[utf8]{inputenc}

\usepackage{microtype}

\newcommand{\genshort}{\textsc{generic (short)}}
\newcommand{\genlong}{\textsc{generic (long)}}
\newcommand{\guideshort}{\textsc{guideline (short)}}
\newcommand{\guidelong}{\textsc{guideline (long)}}
\newcommand{\dd}{\textsc{data-driven}}

\title{Identifying Fairness Issues in Automatically Generated Testing Content}

\author{Kevin Stowe$^1$, Benny Longwill$^1$, Alyssa Francis$^1$ \\
\textbf{Tatsuya Aoyama$^2$\thanks{* Work done while at ETS}, Debanjan Ghosh$^1$, Swapna Somasundaran$^1$} \\ 
$^1$Educational Testing Service (ETS), Princeton, New Jersey \\ 
$^2$Georgetown University}

\begin{document}
\maketitle
\begin{abstract}
Natural language generation tools are powerful and effective for generating content. However, language models are known to display bias and fairness issues, making them impractical to deploy for many use cases. We here focus on how fairness issues impact automatically generated test content, which can have stringent requirements to ensure the test measures only what it was intended to measure. Specifically, we review test content generated for a large-scale standardized English proficiency test with the goal of identifying content that only pertains to a certain subset of the test population as well as content that has the potential to be upsetting or distracting to some test takers. Issues like these could inadvertently impact a test taker's score and thus should be avoided. This kind of content does not reflect the more commonly-acknowledged biases, making it challenging even for modern models that contain safeguards. We build a dataset of 601 generated texts annotated for fairness and explore a variety of methods for classification: fine-tuning, topic-based classification, and prompting, including few-shot and self-correcting prompts. We find that combining prompt self-correction and few-shot learning performs best, yielding an F1 score of 0.79 on our held-out test set, while much smaller BERT- and topic-based models have competitive performance on out-of-domain data.\footnote{Code and dataset available at \url{https://github.com/EducationalTestingService/fairness-detection}.}
\end{abstract}

\input{sections/introduction}    
\input{sections/related_work}

\input{sections/motivation}

\input{sections/methods}
\input{sections/results}
\input{sections/conclusions}
\input{sections/ethics}
\input{sections/limitations}

\bibliography{anthology,custom}
\bibliographystyle{acl_natbib}

\appendix

\input{sections/appendix}

\end{document}

%% file: sections/introduction.tex
\section{Introduction}

    \begin{figure}[t]
    \small
    \includegraphics[width=.48\textwidth]{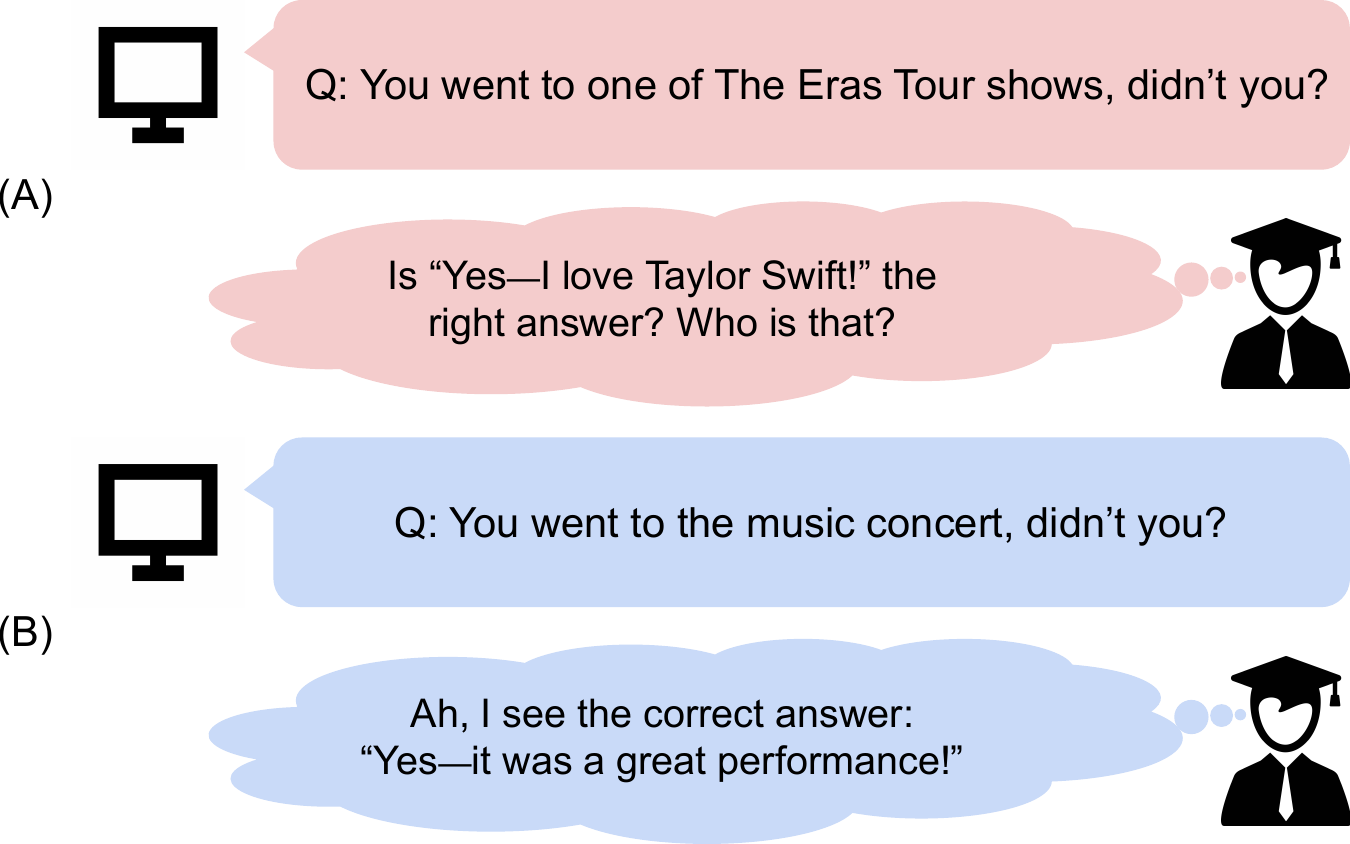}
    \centering
    \vspace{-.5em}
    \caption{\label{fig:taylor} In (A), the generated question requires knowledge of what The Eras Tour is to identify the correct answer. Even native English speakers would likely not be able to identify the correct response if they were not familiar with Taylor Swift. In (B), the generated question does not require specific background knowledge, so test takers would not need to use specialized knowledge to identify the correct answer. Our goal is to identify and filter content like (A) to help ensure fair testing.}
    \vspace{-1.5em}
    \end{figure}

    Large language models (LLMs) have become ubiquitous in the space of natural language generation (NLG) due to recent advances in model capability \cite{minaee-2024}. However, these improvements come with the potential for various negative societal impacts. These negative impacts include the generation of misinformation/propaganda, allocation harms of systems providing benefits only to certain groups of people, and representational harms revolving around bias and stereotyping. Natural language processing (NLP) models--including LLMs--are known to reflect and repeat harmful biases and stereotypes \cite{hosseini-2023,bender-2021,hovy-2021,nadeem-2021}, and research into how the community addresses the societal harms engendered by NLP technology is critical \cite{wang-2024,dev-2022,blodgett-2020}. 
    
    Many of these types of bias in language generation are well-studied. Biases based on gender \cite{nemani-2024,devinney-2022,strengers-2020,wan-2023}, race \cite{das-2022,field-2021}, nationality \cite{venkit-2023}, and disability \cite{venkit-2022} have been identified in language models, and many modern LLMs incorporate deliberate safeguarding measures in an attempt to alleviate these issues \cite{openai-2023,anil-2023}. 
    
    In the area of language assessment, there exists a tangential set of issues regarding fairness to test takers and score users \cite{ets}.  These issues are particularly dangerous when applied to language learning and assessment; tests with inherent biases have the potential to compromise the validity of the test. Therefore, content that is irrelevant to the skills and abilities the test is intended to measure should be avoided (Figure \ref{fig:taylor}). This includes content that could disadvantage anyone based on their culture, location, or experiences (e.g., focusing on barbeques on the 4th of July could disadvantage test-takers who are unfamiliar with U.S. culture); their emotions (e.g., health hazards and diseases can evoke negative emotional responses among some people); their worldviews (e.g., luxury cruises or designer clothing may make some people feel excluded); and other factors. We refer to these types of issues as \textbf{fairness} issues. Knowing how to better understand, detect, and mitigate bias related to fairness in NLG not only raises awareness of the issue but also enables researchers and developers to create more fair and inclusive NLP systems, evaluation metrics, and datasets in the language assessment space.  
    %The items in assessments should not disadvantage anyone based on their culture, location, native language, or expertise with particular topics: these issues appear benign in other contexts but can be harmful in the assessment context. We refer to these issues as \textbf{fairness} issues. 
    
    %Fair testing requires items that do not contain irrelevant factors that negatively impact the assessment of a test taker. These factors include questions focusing on specific aspects of one culture, location, or topic (eg. focusing on barbeques, 4th of July unfairly advantages test takers from the United States); evoking negative emotional responses with controversial/triggering content, such as health hazards and diseases; questions about topics that someone of a particular economic class is unlikely to be familiar with (eg. luxury cruises); and other issues. Knowing how to better understand, detect, and mitigate bias related to fairness in LLM-NLG not only raises awareness of the issue but also enables researchers and developers to create more fair and inclusive NLP systems, evaluation metrics, and datasets in the language assessment space.  

    Our goal is to build a system for identifying fairness-violating content in automatically generated texts. It is of course still necessary to have human review and revision of the content, but by adding a filtering process after generation and before manual review, we can significantly reduce the time taken for reviewing and the chance that fairness-related content is mistakenly allowed. To accomplish this goal, we explore four different approaches: fine-tuning, topic-based classification, few-shot prompting, and prompt-self correction.

    Our methods need to adapt to new contexts: our definition of fairness is operationally defined by the particular testing context, and may not apply to others, so the guidelines, prompts, and models may not apply generally to new contexts. For this reason, we assess our methods on two held-out test sets and analyze how our methods could be applied to new contexts. We release our resulting dataset, consisting of 620 samples, of which 19.4\% contain fairness issues\footnote{Each sample we used was rejected for deployment in actual tests. Using rejected samples for our experiments allows us to release the dataset: accepted stimuli cannot be made public.}, to facilitate improvements in the fairness-detection community.

    Our contribution consists of the following:

    \begin{enumerate}
    \itemsep 0em
        \item We define a new fairness problem around issues faced in developing fair testing content.
        \item We release a dataset of 601 samples for use in evaluating fairness detection methods.
        \item We analyze the relative effectiveness of a variety of well-known classification techniques.
        \item We provide a new mechanism for prompting self-correction, which yields significant improvements over other prompting strategies.
    \end{enumerate}

    We start with data collection and analysis. We collect 620 samples over seven different types of content generated using LLM prompting. We annotate each sample and assess whether it contains a fairness issue, and if it does, whether that fairness issue pertains to \textit{knowledge, skill, or expertise} or \textit{emotion} (more on these categories and how they relate to fairness in Section \ref{sec:data}). We then use this dataset to experiment with a series of models for classifying fairness issues.

    We show that fine-tuning and filtering by topic can be cheap and effective options, although prompting strategies with GPT4 tend to be more effective. Few-shot prompting along with self-correcting prompt strategies yield strong performance with relatively little data, and combining both yields the best results on our in-domain test set, with an F1 score of .773. Interestingly, using a shorter, more generic prompt combined with our self-correction method yields the best result on our out-of-domain test set, with an F1 score of .462.

%% file: sections/related_work.tex
\section{Related Work}
    Bias, fairness, and responsible AI has been at the forefront of education technology, with contemporary research focusing on automated scoring, writing assistance, and other nuances of applying NLP technology to this sensitive domain \cite{mayfield-2019,loukina-2019}.   
    \newcite{baffour-2023} find that assisted writing tools may exhibit moderate bias depending on the task, while \newcite{wambsganss-2023} found no significant gender bias difference in writing done with and without automated assistance. \newcite{wambsganss-2022} explore bias in educational tools for German peer review, and \newcite{kwako-2023,kwako-2022} propose novel methods for detecting bias in automated scoring algorithms.

    We are specifically interested in applications to language generation, and there is also substantial work in using LLMs and other NLP technology to generate content for educational assessments \cite{laverghetta-2023,gonzalez-2023,heck-2023,uto-2023,tack-2023,stowe-2022}. However, this work largely fails to address bias and fairness issues in content generation. Our work is specifically focused on fairness issues in automatically generated language testing content.

    In the context of language models, fairness and bias have emerged as critical concerns. Existing detection and mitigation tools generally diverge from our work: some are overly domain-specific like the focus on news articles in \citet{raza2022dbias}, while others are focused on assessing issues within the language models and datasets \cite{aif360-oct-2018}, rather than the outputs. Other works rely on retrospective metrics that assess a model's fairness through aggregated predictions and subgroup analysis, and/or focus on classification rather than generation problems \citep{weerts2023fairlearn,wisniewski-biecek-2022,saleiro2019aequitas}. Although these tools enhance transparency and accountability for evaluating language model issues, they fundamentally differ from our bias detection approach tailored for evaluating generated text in real-time for a production environment.

%% file: sections/motivation.tex
\section{Problem Motivation}
    \label{sec:data}
    In the language testing context, we face a unique set of fairness challenges in generating content. Specifically, fair testing requires content that does not contain irrelevant factors that negatively impact the assessment of a test taker. 

    A primary concern is to ensure that the test content measures only what it is intended to measure. For English-language proficiency tests, this means that the test must measure only the skills and abilities needed to communicate effectively in English, and not other constructs such as background knowledge of specific jobs, events, or cultures.  
    
    Consider the following question and an example of a response to that question: 
    
    \vspace{-.25em}
    \begin{itemize}
    \itemsep -.225em 
    \item Question: You went to one of The Eras Tour shows, didn't you?
    \item Response: Yes--I love Taylor Swift! 
    \end{itemize}
    \vspace{-.25em}
    
    If the task were to identify whether the response is an appropriate response to the question, even some native English speakers would likely get it wrong. This is because, in addition to needing to know features of English proficiency (in this case, the ability to infer gist, purpose, and basic context based on information stated in short spoken texts), one would also need to know about Taylor Swift and her concert tour. Thus, those familiar with Taylor Swift would have an unfair advantage in identifying the correct answer.
    
    Eliminating the fairness issue for this type of question would result in the following revision: 

    \vspace{-.25em}
    \begin{itemize}
        \itemsep -.225em  
        \item Question: You went to the music concert, didn't you?
        \item Response: Yes--it was a great performance!
    \end{itemize} 
    \vspace{-.25em}

    In addition to avoiding testing outside knowledge, it is also important that language proficiency tests do not include content that is offensive or disturbing. For example, the following question and response refer to serious health issues, which have the potential to evoke deep negative emotions. 

     \vspace{-.25em}
    \begin{itemize}
        \itemsep -.225em
        \item Question: Did you hear that Luis has been hospitalized?  
        \item Response: No, but I knew he had a bad case of Covid-19. 
    \end{itemize}
      \vspace{-.25em}
    
    Content like this that could prompt strong feelings of anger, sadness, or anxiety should be avoided because it could derail a test taker’s concentration, resulting in lower performance on the test. How a test taker interacts with this test content may tell more about their ability to concentrate under emotional strain than about their ability to identify a response's linguistic appropriateness. Eliminating this construct-irrelevant content helps to ensure that the test measures only the skills and abilities it is intended to measure. 
    %Therefore, construct-irrelevant topics that a group might find offensive or disturbing should be avoided, because they could result in test takers answering questions incorrectly or inaccurately because they are distracted. 
    
    %In sum, eliminating content with fairness issues from assessments ensures that only the test’s intended constructs are measured. Critically, these issues are unlikely to be flagged as harmful by standard LLM evaluations: content generated within certain topics that may be unfair to test takers is perfectly normal, and neutral content in other contexts. Any safeguards, either public or privately implemented by the owners of proprietary LLMs, are unlikely to detect or handle the issues in this context.

%% file: sections/methods.tex
\section{Methods}
    \label{sec:methods}

    Our goal is to detect whether a generated stimulus contains an issue as a binary classification task. We build a dataset of texts labeled for potential fairness issues and explore potential detection methods.
    
    \subsection{Dataset}

    \begingroup
    \setlength{\tabcolsep}{4pt} % Default value: 6pt
    \renewcommand{\arraystretch}{1.1} % Default value: 1
 \begin{table}
        \small
        \centering
        \begin{tabular}{c|c|c|c|c}
            \textbf{Item/Task Type} & \textbf{Total} & \textbf{Fairness} & \textbf{KSA} & \textbf{Emotion}  \\
            \hline
            Read a Text Aloud & 304 & 55 & 24 & 39 \\ 
            \hline
            Talks & 91 & 12 & 6 & 6 \\
            Text completion & 84 & 26 & 11 & 19 \\
            \hline
            \makecell{Respond to \\ Questions Using \\ Information Provided}  & 56 & 10 & 5 & 5 \\
            \hline
            \hline
            {*Conversations} & 41 & 8 & 5 & 4 \\
            \hline
            \makecell{*Respond to a \\ Written Request} & 25 & 7 & 6 & 1 \\
 %           \hline
 %           {*Announcements} & 19 & 3 & 1 & 2 \\
            \hline
            \hline
            Total & 601 & 118 & 57 & 74 \\
        \end{tabular}
        \caption{Item/task types and annotations for fairness issues. Each has a binary annotation (fairness issue/no fairness issue) and is tagged as containing a KSA issue or an Emotion issue. Types marked with '*' are held out for testing as an "out-of-domain" dataset, and not used for any training/evaluation.}
        \label{tab:itemcount}
    \end{table}
    \endgroup
    
    Our goal is to identify and mitigate these fairness issues in testing content. We build a dataset spanning seven different item or task types from standardized English language proficiency tests all generated using GPT4 \cite{openai-2023}. Item and task types can contain up to four components: the stimulus (main text the question is based on), stem (question asked about the stimulus), key (the correct answer to the stem), and distractors (a set of alternative answers that are incorrect).
    
    Fairness issues are possible in all components, but we focus on only the stimuli, which are typically the longest, most feature-rich components of the test content, and thus are most likely to reflect fairness and bias issues. Issues in the stimuli can leak through to other components, making the stimulus the source of the majority of fairness issues.

    \paragraph{Annotation}
    For each stimulus, we aim to identify whether or not the stimulus contains fairness/bias issues, and if so, what type of issue is present. We start with a dataset of automatically generated stimuli. These stimuli were generated using prompting and different versions of GPT: the prompts were iteratively improved with the goal of improving the overall quality of the stimuli. During this process, each stimulus was evaluated by the test's content development experts. For this work, the stimuli used were rejected by the reviewers, allowing us to provide them publicly and explore their use for fairness detection. These rejected stimuli typically have the relevant language and structure, so our goal is to identify which of those stimuli were rejected (at least in part) for fairness reasons. We employ content development experts to annotate these samples, yielding a binary classification between non-fairness and fairness-related rejections.

    However, there are different ways for bias and fairness considerations to impact individual stimuli. To better understand and mitigate these issues, we separated them into two main categories:
    
    \begin{itemize}
        \itemsep -.15em
        \item \textit{Knowledge, Skill, and Ability (KSA)}: content that contains construct-irrelevant information that may be unavailable to test takers in different environments or with different experiences and abilities. These include content with reference to specific skills, regionalisms, or unfamiliar contexts.
        \item \textit{Emotion}: content in which language, scenarios, or images are likely to cause strong emotions that may interfere with the ability of some groups of test takers to respond. These include offensive, controversial, upsetting, or overly negative content.
    \end{itemize}
    
    Each sample that is flagged for fairness is annotated for one or both of these categories. This allows further analysis to address these specific fairness categories and to better understand the impact of specific fairness issues.

    Our dataset is comprised of stimuli from seven different item and task types: a summary of the collected data is shown in Table \ref{tab:itemcount}, with examples for each type in Appendix \ref{app:items}. These stimuli represent various structures, depending on the item/task type: Read a Text Aloud, Talks, and Text Completion stimuli are short text paragraphs, while Conversation stimuli involve turns between two or more speakers. Respond to Questions Using Information Provided and Respond to a Written Request task stimuli are structured content: the generation process creates text that is filled into a structured template; we use only the raw text.
    
    Overall we collect 601 samples, of which 19.6\% exhibit evidence of fairness issues, with 9.5\% reflecting KSA issues and 12.3\% Emotion issues. We build a validation set of 48 samples reflecting a balance of the item and task types from the training types (Read a Text Aloud, Talks, Text Completion, and Respond to Questions Using Information Provided), and an equal-sized "in-domain" dataset from these stimuli is held separately for testing. These datasets contain an even number of positive and negative classes for fairness evaluations. As our goal is to be able to identify positive cases where fairness issues exist, we intend for our validation and test sets to have a substantial number of this class. We use the two remaining types (Conversations, Respond to a Written Request) as a separate "out-of-domain" test set to evaluate performance on unseen content.

    \subsection{Experiments}
    We experiment with standard transformer-based classification baselines, topic detection, and a variety of GPT4-based prompting, including methods for automatic prompt-self correction. We describe each method below: each is tuned on the validation set, and we report the best model performance on that set. We then evaluate model performance on two separate test sets in Section \ref{sec:results}.

    \paragraph{Classification with Fine-Tuning}
    We fine-tune standard pre-trained transformer models for sequence classification. We experiment with \texttt{bert-base-cased}, \texttt{bert-large-cased} \cite{devlin-2019},  \texttt{roberta-base}, \cite{liu-2019} and \texttt{deberta-base} \cite{he-2021} models. We perform a hyperparameter search on our validation set for each model, finding that a learning rate of \texttt{2e-5} over 2-4 epochs generally performs best, and report results using the model with the best performance.

    \paragraph{Topic-Based Filtering}
    We observe that many samples are flagged for fairness due to the topic of the material: many topics contain content that violates our fairness guidelines directly, while others are simply more likely to include unacceptable content. Motivated by this, we explore topic detection as a method for identifying fairness issues.

    We first identify topics found within the data. We use the topic modeling framework BERTopic \cite{grootendorst2022bertopic} to extract topic representations from two sources of training data: (1) all samples from the training partition of our dataset and (2) our fairness guidelines. In this method, SentBERT \cite{reimers-2019-sentence-bert} converts each training document into a dense vector representation which are then grouped by semantic similarity, creating clusters that represent different topics. For each of the two training sets, topic descriptions made up of the most important words in a cluster are generated for the clusters containing at least five supporting documents. We manually assess each topic description for themes that should be avoided based on their relation to known fairness issues and which topics are acceptable. Finally, for each unseen sample in test and validation datasets, we make predictions based on the single nearest topic cluster. If a sample falls within the boundaries of restricted topics, it is classified as a violation.

    \begin{table}[t]
        \centering
        \small
        \begin{tabular}{c|c|c|c}
            \multicolumn{4}{c}{\textbf{Fine-tuning}} \\
            Model & Prec & Rec & F1 \\
            \hline
            \texttt{bert-base-cased} & 1.00 & 0.29 & 0.45 \\
            \texttt{bert-large-cased} & 0.92 & 0.50 & 0.65  \\
            \texttt{roberta-base} & 0.92 & 0.50 & 0.65  \\
            \texttt{deberta-base} & 1.00 & 0.63 & 0.77 \\
        \end{tabular}
        \vspace{.5em}
        
        \begin{tabular}{c|c|c|c}
        \multicolumn{4}{c}{\textbf{Topic-based Filtering}} \\
            Model & Prec & Rec & F1 \\
            \hline
            Topic-data & 0.79 & 0.46 &	0.58  \\
            Topic-guidelines & 1.00 & 0.04 & 0.10
        \end{tabular}
        
        \caption{\label{tab:firstresults}Results for fine-tuning (above) and topic detection (below) on the validation set.}
    \end{table}

    Results for these methods are shown in Table \ref{tab:firstresults}. The fine-tuned bert-based models perform fairly well, with F1 scores for \texttt{bert-large-cased} and \texttt{roberta-base} both around 0.65, and \texttt{deberta-base} showing exceptional performance with an F1 score of 0.77. The Topic-Based Filtering models are worse, with the data-based system yielding an F1 score of 0.58. In all cases, precision is much higher than recall; these models are conservative with predictions.

    \subsection{Prompting}
    
    We initially experiment with five different ``base'' prompts. We pair these with stimuli and use GPT4 to return ``True'' if the stimulus contains a fairness issue and ``False'' otherwise. These prompts represent different strategies\footnote{Prompts in Appendix \ref{app:prompts}.}:
    
    \begin{itemize}
    \itemsep -.15em
        \item \textbf{\genshort\, 53 tokens}: Drawing from general knowledge of fairness and bias in LLMs, we write a generic prompt designed to combat attested LLM biases. This prompt is designed as a weak baseline. Our goal is to determine if a short, simple prompt can capture relevant issues, and whether or not it can be easily improved via self-correction or few-shot learning (Sections \ref{sec:fewshot} and \ref{sec:correction})
        \item \textbf{\genlong\, 191 tokens}: This is a longer, more detailed version of the above, containing nearly 200 tokens.
        \item \textbf{\guideshort\, 197 tokens}: We craft a prompt based on guidelines for writing fair assessments. Using documentation that defines what constitutes fair assessment items and how to write them, we build a prompt capturing the important components of a fair question. The goal of this prompt is to determine whether human-written guidelines based on theoretical issues will accurately capture these issues in real data.
        \item \textbf{\guidelong\, 1081 tokens}: We construct a ``long'' version of the previous guidelines by summarizing the entire fairness guidelines with the help of GPT4, asking for concise versions of relevant sections and combining them into a document that fully captures all the relevant aspects of the guidelines. This prompt is our longest, but still fully based on documentation. The goal of this prompt is to determine the efficacy of a longer, more comprehensive prompt.
        \item \textbf{\dd\, 142 tokens}: We craft a prompt based on annotations in our data. We identify which topics and language cause fairness issues and build the prompt to reflect how they might generalize to unseen item/task types and topics. This method is hypothesized to be the most effective, as it will address known issues in the data but may not extend to unseen data, as it is built specifically around the given training samples.
    \end{itemize}

    These prompts are run through GPT4 via the Azure interface \cite{openai-2023}. Each prompt was updated manually to correct obvious potential issues. Our goal here is not to overoptimize prompt writing, which could lead to overfitting the validation set, but rather to develop a generic prompt likely to be effective for both known fairness issues and novel issues possible in generated content.

    Initial experiments on the validation set revealed two insights: the \genlong\ prompt performs similarly to the \genshort\ in all cases, and the \guidelong\ prompt is ineffective. We therefore focus our efforts on the three other prompts: \genshort\, \guideshort\, and \dd.

    A primary issue in identifying the fairness issues we are concerned with is that they are only applicable in the context of certain testing environments. The guidelines and data we use are specific to a certain test; for other tests, other environments, and other contexts, different definitions of fairness and different alignments will be applicable. For this reason, the ability to adapt to new environments is critical in applying fairness constraints. We aim to build a system that can learn to adapt to new guidelines with minimal information.

    We explore several methods to improve the robustness of prompting: few-shot prompting and self-correcting prompting techniques. 

    \paragraph{Few-Shot Prompting}
    \label{sec:fewshot}
    We follow few-shot prompting strategies that have been successful for LLM classification tasks \cite{logan-2022,brown-2020}. For each of our base prompts, we provide a set number $n$ of additional samples for both the positive and negative classes. Each of the samples contains the text of the stimulus along with a brief description of why it does or does not violate our fairness guidelines.

    \paragraph{Prompt Self-Correction}
    \label{sec:correction}
    
\begin{figure}[t]
    \small
    \includegraphics[width=.5\textwidth]{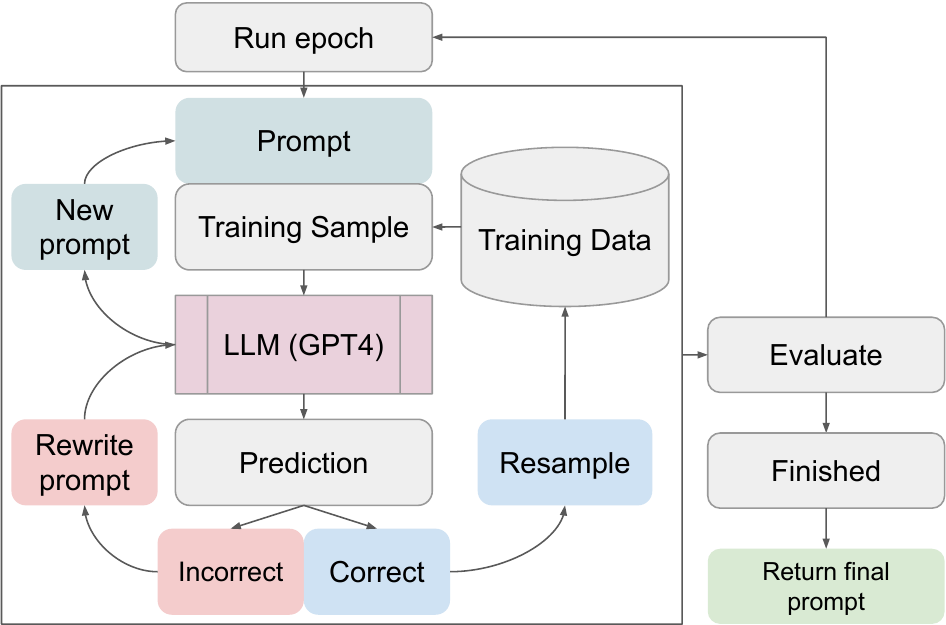}
    \centering
    \vspace{-.5em}
    \caption{\label{fig:selfcorrection} Self-correcting prompt strategy. Data is run through the prompt. If the result is correct, we continue; otherwise, we instruct the LLM to correct the prompt.}
    \end{figure}

    We leverage our annotated data using a prompt self-correction strategy (Figure \ref{fig:selfcorrection}). We start with a base prompt and iterate through a batch of samples from our training data. For each sample, we insert text into the prompt, perform a pass through GPT4, and evaluate the resulting output. If it is correct, we continue. If not, we instruct the LLM to make adjustments to the prompt, either to disallow the sample (when the error is a false negative) or to allow it (when the error is a false positive), using the following prompts:

    \begin{itemize}
        \itemsep -.15em
        \item \textit{Positive sample incorrectly classified as negative}: The previous text was classified as not having any fairness issues, returning False. However, it did have an issue. Think about what topics or issues in the text might be unfair. Improve the original fairness instructions by adding or modifying restrictions so to account for this text. 
        \item \textit{Negative sample incorrectly classified as positive}: The previous text was classified as having a fairness issue, returning True. However, this item has no issue. Think about why this text was classified as unfair. Improve the original fairness instruction to allow this item by removing or revising restrictions.
    \end{itemize}
    
    This process is run up to $e$ epochs, or stopped early if accuracy reaches 1 or the predictions are stable. We run over $b$ batches of $n$ samples randomly drawn from the training data, using the best-scoring prompt from the final batch for evaluation.\footnote{For an example of the process, see Appendix \ref{app:selfcorrection}.}

    \paragraph{Combining Few-Shot and Self-Correction}
    Few-shot and self-correction are inherently complimentary, as the self-correction method returns an optimized prompt and few-shot learning reinforces it by providing in-domain examples. We combine them by concatenating additional few-shot learning samples to the self-correcting prompts.

    For each of these improvements to prompting, we perform a hyperparameter search over the number of total training/few-shot samples and batch size. We experiment with the \genshort\, \guideshort\, and \dd\ prompts.\footnote{Experiments with the longer guideline-based prompt were unsuccessful: the LLM invariably returns either a commentary on a single testing procedure or rewrites the prompt entirely to handle a single sample.} We hypothesize the \genshort\ and \guideshort\ prompts should be able to benefit quickly from adaptive methods, while the \dd\ prompt should be nearly optimized, as it is already based on observations from the data.

\begin{figure}[t]
    \small
    \includegraphics[width=.5\textwidth]{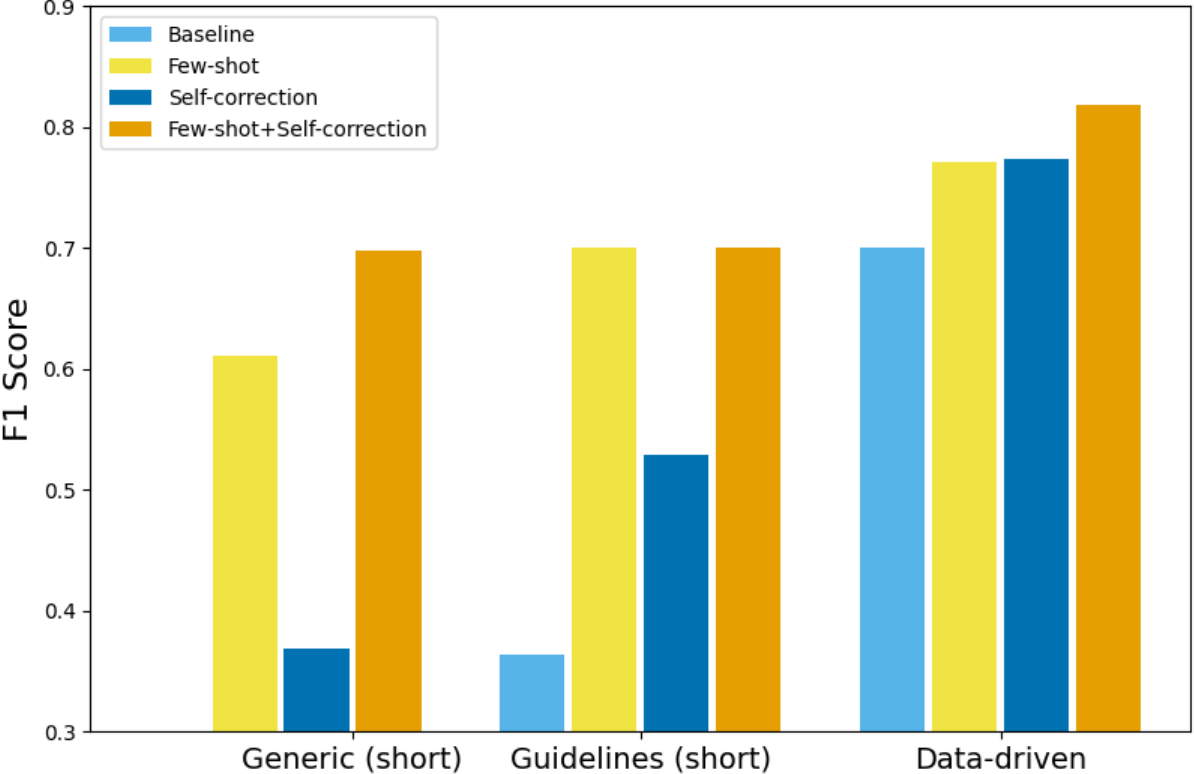}
    \centering
    \vspace{-.5em}
    \caption{\label{fig:prompting} F1 scores on the validation set for each prompting method. Note that for \genshort\, the F1 score was 0. Full results in Appendix \ref{app:valid}.}
    \vspace{-1.5em}
    \end{figure}

    We use the validation set to tune the prompts and parameters to optimize the F1 score for each method. Note that for all prompting strategies, the temperature is set to zero; the prompts should only return True or False. Figure \ref{fig:prompting} shows the best results on the validation set. We explore each model's effectiveness on unseen data in Section \ref{sec:results}.
    
    The base generic prompt fails, as the traditional bias and stereotyping issues are less likely to occur in our generated content, and the fairness issues we are concerned with are unlikely to be deemed as problematic out of context. Using a simplified version of our guidelines yields a 0.36 F1 score for identifying fairness issues.  The \dd\ based on observations in the training data yields much better results (0.70 F1). However, this may not extend well to novel cases, as the prompt is driven purely by our validation data.
    
    Few-shot learning displays some interesting properties: we see significant improvements across all three prompts, using three samples. (This yielded the best results across all validation runs). Even the minimal \genshort\ prompt rises to over 0.60 F1 with minimal few-shot prompting. 
        
    We see small improvements over the baseline using prompt self-correction for all three prompts. For the \dd\ prompt, results using self-correction equal those using few-shot learning. This aligns with previous work showing that language models themselves tend to write better prompts \cite{fernando-2023}: after only a few iterations of self-correction, the \dd\ prompt surpasses the performance of a human-written prompt, even in cases where the human describes the dataset explicitly.

    Combining self-correction and few-shot learning yields improvements over base prompts and few-shot prompting alone. This approach yields the best results for all three prompts, with the best-performing model being the \dd\ prompt with self-correction and few-shot learning. This may be due to overfitting, however: the prompt is written to reflect the data. To explore the efficacy of these methods on unseen data, we evaluate them on our two held-out test sets.

%% file: sections/results.tex
\section{Test Results}
\label{sec:results}
    
    \begin{figure*}[t] 
    \small
    \includegraphics[width=\textwidth]{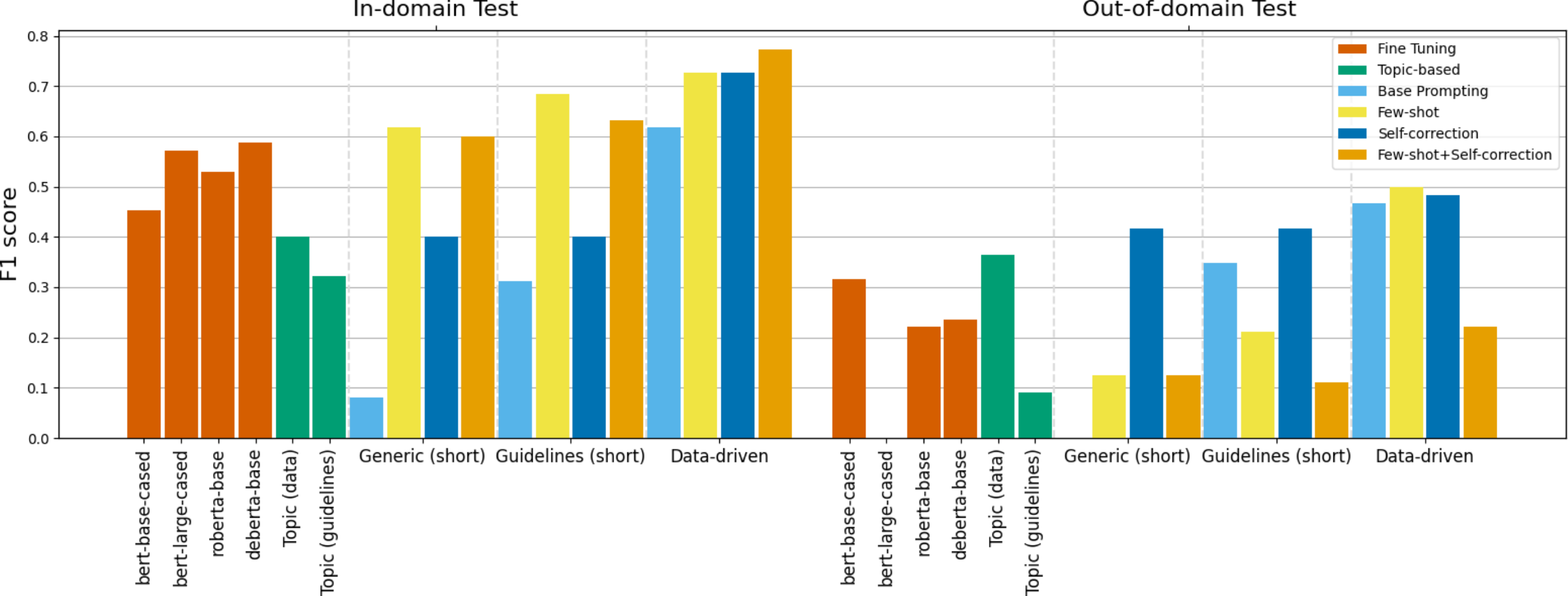}
    \centering
    \vspace{-.5em}
    \caption{\label{fig:final} F1 scores on two test sets for each proposed method. Note that for \texttt{bert-large-cased} and \genshort, the scores were 0.00 on the unknown test set. Full results in Appendix \ref{app:test}.}
    \vspace{-1.5em}
    \end{figure*}

The previous experiments describe our attempts to identify the best-performing model for fairness classification on our validation set. Our goal is to develop a system that generalizes. For this, we evaluate the best-performing of the above model types on two held-out test sets:

\begin{enumerate}
\itemsep -.15em
    \item \textbf{In-domain:} The 48 held-out samples drawn from the item/task types used for training.
    \item \textbf{Out-of-domain:} All samples (66) from the two held-out types: Conversations, Respond to a Written Request.
\end{enumerate}

    Figure \ref{fig:final} shows the results on the test set. We evaluate the best-performing models of each type: fine-tuned transformer models, topic-based classification, base prompts, few-shot learning, self-correction, and combining few-shot and self-correction. We here note some key facts about model performance on our test set.
    
 \paragraph{Best Performance}
    Combining the \dd\ prompt with self-correction and few-shot learning performs the best on the in-domain test. This shows this is the best approach if there is available data and expertise to support hand-crafting a \dd\ prompt and running self-correction. On the out-of-domain data, the smaller initial prompts, \genshort\ and \guideshort\, both outperform the \dd\ prompt, perhaps due to their more generic nature: the \dd\ prompt is too specific to this dataset, and understandably doesn't generalize well. The self-correct+few-shot methodology performs the best in both cases: few-shot learning alone is better than self-correction alone, but the combination is typically the best.

    \paragraph{Strong Results from Small Models}
    Traditional transformer-based classification performs remarkably well, especially in generalizing to the out-of-domain data. On the in-domain data, the best performing model \texttt{deberta-base} performs on par with the best base prompting model (0.58 compared to 0.60 F1 score), although this is a significant drop from the validation performance of 0.77, and performs quite poorly on out-of-domain data (0.20), indicating the model may overfit during training. On the out-of-domain data, \texttt{roberta-base} performs nearly as well as the best-performing overall model, just 0.04 behind the \genshort\ prompt with self-correction and few-shot learning. If the goal is to quickly and cheaply build a system that is applicable to a wide variety of domains, there appears to be significant value in relying on these relatively small transformer-based classification models. The Topic (data) approach is also competitive on out-of-domain data, and does not even require model training; it lags only slightly behind the \texttt{roberta-base} model.

    \paragraph{Self-Correction}
    We found significant success in our proposed self-correction mechanism. While it typically does not outperform few-shot learning in isolation, the methods are naturally complementary, and the combination often yields the best-performing model. In examining the models' self-corrections, we find that when asked to become more restrictive, the model tends to add sentences with new constraints, which nicely reflect the issue that was missed. When asked to become less restrictive, the model tends to add hedges to currently existing constraints.

    In our experiments, we noted some issues. First, when run using too many samples or batches, the prompts tend to degrade: once the LLM makes an error and returns a prompt that doesn't match the specifications, the run needs to be aborted. Even when the LLM sticks to the instructions, after many iterations the prompts become unwieldy and self-contradictory, and performance rapidly declines. We suggest using somewhere between six and 20 total samples for prompt self-correction; it is best to avoid making corrections indefinitely.

    \paragraph{Use-Cases and Metrics}
    We here report F1 score as a balance between precision and recall. (For full scores, see Appendix \ref{app:test}.) Depending on the end use case, other metrics may be more appropriate. In our case, we advocate for always including humans in the evaluation process to ensure that only fair content is accepted. We then value both precision (as we do not want to excessively flag content for fairness issues, which could reduce diversity) and recall (as we do not want to let fairness issues through). Optimizing for recall seems reasonable, as it is likely more important to prevent fairness issues from being released, but it is critical to note that no system is perfect: even optimizing for recall, these fairness issues are likely to persist, and the models should not be used as failproof safeguards.

      \paragraph{KSA and Emotion}

            \begingroup
    \setlength{\tabcolsep}{3.5pt} % Default value: 6pt
    \renewcommand{\arraystretch}{1.1} % Default value: 1
        \begin{table}[t]
        \centering
        \small
        \begin{tabular}{c|c|c|c}
            \textbf{Model} & \textbf{Type} & \textbf{KSA} & \textbf{Emotion} \\
            \hline
            \multirow{3}{*}{Fine-tuned} & \texttt{bert-base-cased} & 0.07 & 0.57 \\
            & \texttt{bert-large-cased} & 0.00 & 0.00 \\
            & \texttt{roberta-base} & 0.06 & 0.56 \\
            & \texttt{deberta-base} & 0.08 & \textbf{0.75} \\            \hline
            \multirow{2}{*}{Topic-based} & Data & 0.26 & 0.59 \\
            & Guideline-based & 0.20 & 0.06 \\
            \hline
            \multirow{3}{*}{Base Prompting} &  \genshort\ & 0.00 & 0.00 \\
            & \guideshort\ & 0.29 & 0.09 \\
            & \dd\ & \textbf{0.47} & 0.50 \\
            \hline
            \multirow{3}{*}{Self-correction} & \genshort\ & 0.35 & 0.30 \\
            & \guideshort\ & 0.35 & 0.27 \\
            & \dd\ & \textbf{0.47} & 0.41 \\
            \hline
            \multirow{3}{*}{Few-shot} &  \genshort\ & 0.18 & 0.24 \\
            & \guideshort\ & 0.30 & 0.24 \\
            & \dd\ & 0.36 & 0.56 \\
            \hline
            \multirow{3}{*}{\makecell{Few-shot +\\Self-correction}} &  \genshort\ & 0.18 & 0.21 \\
            & \guideshort\ & 0.23 & 0.21 \\
            & \dd\ & 0.24 & 0.59 \\
        \end{tabular}
        \caption{\label{tab:ksa} Recall scores for KSA and Emotion-labeled data across both test sets.}
        
        \end{table}
    \endgroup

    \begin{comment}
            \begin{tabular}{c|c|c|c}
            \textbf{Model} & \textbf{Type} & \textbf{KSA} & \textbf{Emotion} \\
            \hline
            \multirow{3}{*}{Fine-tuned} & \texttt{bert-base-cased} & .000 & .107 \\
            & \texttt{bert-large-cased} & .116 & .107 \\
            & \texttt{roberta-base} & .116 & .320 \\
            \hline
            \multirow{2}{*}{Topic-based} & Data & .289 & \textbf{.571} \\
            & Guideline-based & .182 & .052 \\
            \hline
            \multirow{3}{*}{Base Prompting} &  \genshort\ & .000 & .000 \\
            & \guideshort\ & .226 & .077 \\
            & \dd\ & .381 & .500 \\
            \hline
            \multirow{3}{*}{Self-correction} & \genshort\ & .161 & .235 \\
            & \guideshort\ & .277 & .103 \\
            & \dd\ & .271 & .210 \\
            \hline
            \multirow{3}{*}{Few-shot} &  \genshort\ & .212 & .419 \\
            & \guideshort\ & \textbf{.445} & .313 \\
            & \dd\ & .322 & .552 \\
            \hline
            \multirow{3}{*}{\makecell{Few-shot +\\Self-correction}} &  \genshort\ & .387 & .393 \\
            & \guideshort\ & \textbf{.445} & .287 \\
            & \dd\ & .264 & .470 \\
        \end{tabular}
        \caption{\label{tab:ksa} Recall scores for KSA and Emotion-labeled data across both test sets.}
        
        \end{table}
    \end{comment}
    We evaluate performance on the test set for the two subcategories: Knowledge, Skill, and Ability (KSA) and Emotion (Table \ref{tab:ksa}). The \texttt{deberta-base} model performs exceptionally well on the KSA subcategory, capturing 75\% of the fairness-flagged samples. Data-based methods (the \dd\ prompts (0.59) and Topics from Data (0.59)) also perform well, likely due to the inclusion of negative emotional issues in the text. They perform much worse on KSA classification, although the \dd\ prompts still yield the best performance (0.47): KSA-related issues are especially difficult as they generally involve only specific knowledge, and would not normally be considered fairness issues in other contexts.
    

%% file: sections/conclusions.tex
\section{Conclusions}
    %This work details the creation of a fairness dataset based on automatically generated content and an exploration of methods designed to identify fairness issues in natural language generation. We explore fine-tuning, topic-based detection, and prompting, including novel prompt self-correction, showing that combining this new approach with few-shot learning yields the best performance on unseen data.

    This work delivers four key contributions: an exploration of a novel fairness detection task, a dataset of 601 samples annotated for fairness issues, evaluation of a variety of classification models for this task, including fine-tuning, topic-based approaches, and prompting, and a novel prompting strategy, which, combined with few-shot learning, achieves state-of-the-art performance on the task.

    This work is aimed to explore the space of fairness and bias issues in generated content, especially in the education context. We aim to highlight the difficulties of accounting for fairness, particularly in specific contexts unlikely to be accounted for by traditional model guardrails. As language model usage becomes more prevalent, the need for proper bias and fairness strategies from people training, deploying, and using these models is paramount.

%% file: sections/ethics.tex
\section{Ethics}
Content generation comes with inherent ethical concerns relating to fairness, bias, factuality, and sensitivity. Our work aims to mitigate these issues with regard to fairness, but it is important to consider potential issues that might arise from using LLMs and other NLP technology in generating assessment content. Models may introduce subtle biases against disadvantaged groups, or produce content that appears to be factual, but is not. These are critical failures that need to be accounted for.

In practice, the generation of assessment content requires human intervention: large language model generations are not at the point where they are immune to these negative impacts, and thus for any content that goes into production, a human with relevant expertise needs to evaluate it. The methods we propose support this human intervention, as they can remove obviously offensive content before the human review stage, or assist in human reviews by flagging potentially harmful content.

While our dataset is unlikely to contain any content that is triggering (our framework of fairness is focused on more nuanced contexts), it must be noted that there is potential for it to be used maliciously; for example, by someone designing a system to adapt to and deceive a fairness detection system. In releasing this data, we hope to bring awareness to this issue and better understand the potential negative impacts. Primarily, we stress that any fairness detection system should not be used in isolation or without supervision as a catchall for potential issues.

%% file: sections/limitations.tex
\section{Limitations}
Our work is limited largely by the type of content evaluation and the models used. We focus on a small number of item and task types that fall under very specific fairness constraints: the evaluation of the methods used specifically applies to these items under these constraints. This is apparent in the evaluation on the "unseen" item types in Section \ref{sec:results}. Applying these methods to new item and task types, even those annotated under the same fairness guidelines, yields significantly reduced results. This is evidence that the methods and models we designed work only for the specific contexts in which they are trained and developed.

Similarly, we explore a small space of models and approaches. We use relatively basic prompt strategies; there exist many other approaches and improvements that are likely to be valuable that we do not evaluate. The same is true of fine-tuned models and topic classification. We present relatively basic, well-known strategies to better understand the difficulty of our data, with the understanding that there are substantial improvements that could be applied.

%% file: sections/appendix.tex
\section{Item Types}
\label{app:items}
    Table \ref{tab:app-items} gives examples for each item type.
    
  \begin{table*}
        \small
        \centering
        \newcommand{\wrap}[1]{\parbox{.78\linewidth}{\vspace{1.5mm}#1\vspace{1mm}}}
        \begin{tabular}{c|c}
            \textbf{Item Type} & \textbf{Example}  \\
            \hline
            Read a Text Aloud & \wrap{Welcome to our house hunting service. Our priority is to help find your dream home. We offer tours of houses with gardens, pool facilities, or spacious garages. However, we also tackle paperwork which can be confusing for first-time buyers. Join us, start your journey towards owning a home.} \\
            \hline
            Talks & \wrap{Hello, I'm your local council housing officer. I'm reaching out about our new housing construction plans. We are designing affordable, environmentally friendly homes in our area. For your input on these proposed designs, please complete our quick survey. It won't take more than a few questions. Your opinions are valuable in ensuring these homes meet community needs. Share your thoughts, let's create a better living environment together.}  \\
            \hline
            Text completion & \wrap{ORG\_1 is a locally owned gym that offers a wide range of fitness classes and equipment. We are proud to offer two special deals for our members. The first is a one-month membership for only \$50. This includes unlimited access to all our classes and equipment. The second is a three-month membership for \$125. This includes a free personal training session and a 10\% discount on all additional personal training sessions. Come join us today and take advantage of these great deals!}\\
            \hline
            \makecell{Respond to Questions \\ Using Information \\ Provided} & \wrap{N: Hello. I received an email about the annual conference that the Association of Professional Journalists is hosting, but I can't seem to find it. I was hoping you could answer a few questions. \\ Header: ['Annual Small Businesses Conference', 'Riverside Convention Center', 'Daily rate: \$70] \\ Event table: ['', 'Day 1, 9:00 A.M.', '10:30 A.M.', '12:00 P.M.', '2:00 P.M.', 'Day 2, 10:30 A.M.', 'Workshop: Basics of Data Security', 'Speech: Role of Entrepreneurship', 'Lunch Break', 'Presentation: Advanced Data Protection', 'Speech: Customer Relations and Service', 'Hans Pham', 'Hans Pham', '', 'Olga Gomez']} \\
              \hline
            {*Conversations} & \wrap{(Woman) Good morning, Alex. Have you reviewed the department's salary structure for the upcoming year? \\ (Man) Morning, Priya. Yes, I have. There's a significant disparity, especially when considering the recent promotions.\\ (Woman) It seems like those promoted are receiving disproportionately higher raises. \\ (Man) True. I'll propose a more balanced salary increase for all levels at the next management meeting. It should help in retaining our skilled employees.} \\
            \hline
            {\makecell{*Respond to a \\ Written Request}} & \wrap{date: ['January 8'] signature\_line: ['Best regards, Shani Flores, Airline Customer Support Specialist'] from: ['Shani Flores, Airline Customer Support Specialist'] to: ['Hassan Kumar'] body: ['Thank you for choosing our airline for your recent travel.Your feedback is valuable. Could you share your thoughts on what we could do better during the flight?'] subject: ['Improving in-flight services']} \\
            %\hline
            %{*Announcements} & \wrap{Ladies and gentlemen, this is John Smith from Campus Administration. We would like to remind you of our noise policies. Please refrain from loud activities in study areas and respect quiet hours in residential buildings. Violators may face disciplinary action. Thank you for your cooperation.} \\
        \end{tabular}
        \caption{Item types and annotations for fairness issues. Each has a binary annotation for containing a fairness issue and then can be also tagged as containing a KSA issue or an Emotion issue. Multiple subcategories are allowed. *Items are held out for testing, and not used for any training/validation.}
        \label{tab:app-items}
    \end{table*}

\section{Prompts}
\label{app:prompts}

Below are the five base prompts used. Each also contains a slot to insert a stimulus and a suffix instruction: "Consider the instructions. If the text violates the instructions, return only the word True. If not, return only the word False."

\subsection{\genshort\ }
Ensure it is fair to a test taker. It should not contain biased, toxic, or unethical content. It should not reflect any special skills, abilities, or culture, and it should not contain any negative emotion.

\subsection{\genlong\ }
Develop a set of criteria to filter out negative test items that may violate fairness, bias, or sensitivity requirements. Consider the following aspects:

Fairness:
Are the test items equally applicable and appropriate for all individuals regardless of their background, demographics, or personal characteristics?
Do the test items avoid favoring or disadvantaging any specific group based on race, ethnicity, gender, religion, sexual orientation, socioeconomic status, disability, or any other protected characteristic?
            
Bias:
Do the test items avoid stereotypes or assumptions that may lead to unfair advantages or disadvantages for certain groups?
Are the language, examples, scenarios, and references used in the test items neutral and inclusive?
Have the test items been reviewed to ensure they do not unintentionally favor or discriminate against any particular group?

Sensitivity:
Are the test items respectful and considerate of individuals' diverse experiences, perspectives, and identities?
Do the test items avoid potentially triggering or offensive content?

\subsection{\guideshort }
Consider the following text. The text should avoid any content or information that would unfairly disadvantage someone tested on it. It should not do any of the following: 
            
1. It should not contain content specific to a region, including brand names, celebrities, sports, entertainment, culture, customs, geography, government, history, holidays, or laws. Any of these topics that are specific to a certain region should be excluded.

2. It should not include topics that require specialized knowledge that would be easier for one group than another. 

3. It should not include unnecessarily difficult language including words, phrases, and concepts more likely to be known by one group than another. 

4. It should avoid contexts that not all people may have experienced. The contexts should not require direct, personal experience to understand. 

5. It should not mention religion. 

6. It should not include contemptuous, derogatory, or exclusionary language. It should not induce any negative emotions. 

7. It should not advocate for particular causes or ideologies, or include anything divisive. 

8. It should avoid sensitive and controversial topics, including political issues, natural disasters, accidents, or other negative topics.

\subsection{\guidelong }
Below are a set of guidelines. These guidelines aim to enhance the fairness and validity of tests, communications, and other materials. These guidelines assist users in understanding fairness in assessment, including the right content, eliminating unfair content, promoting diversity and inclusivity, addressing accessibility and inclusion issues and reducing subjective fairness decisions. The guidelines cover the fairness of various subjects including the National Assessment of Educational Progress (NAEP), K–12 tests, artificial intelligence (AI) algorithms and includes information to help use plain language and a quick reference guideline list.

  Understanding fairness in testing is crucial for proper application of guidelines, though its definition varies. One common definition sees fairness as absence of any inequity, affecting individuals and groups alike, such as unfair test questions or biased content affecting diverse groups. Another definition argues that tests seeming harder for certain groups aren't necessarily unfair, as differences in results may reflect real differences in knowledge or ability, not test bias. Group score differences don't prove bias, but should be explored to rule out bias. Furthermore, fairness definitions based on outcomes are contested and of limited use during test design. Fairness is also defined based on test validity. The test validity indicates quality, and represents the accuracy of inferences and actions based on scores, which must be equally valid for all test-takers for a test to be fair. Therefore, an effective definition of fairness in assessment is rooted in validity, creating an interconnected relationship between the two. Lastly, fairness in testing relates to the effectiveness of related educational products and services in fulfilling their intended purposes.
  
  These guidelines should ideally cater to everyone, particularly focusing on groups discriminated against due to factors such as age, appearance, citizenship, disability, ethnicity, gender, national origin, native language, race, religion, sexual orientation, and socioeconomic status. It's crucial to also account for intersectionality, a framework recognizing how overlapping identities like race and gender can impact the experiences of individuals with multiple marginalized identities. For instance, Black women may perceive test material differently than Black men or White women.
  
  Principles and Guidelines for Fairness 
  
  Fairness in assessment requires adherence to key principles: Tests should focus on essential aspects of the intended construct and avoid construct-irrelevant hurdles. They must offer design, content, and conditions facilitating valid inferences about diverse test takers' knowledge and abilities. Also, they should provide scores that allow valid group-wise inferences. The subsequent sections offer specific guidelines related to these principles. In case of interpretational conflicts, choose the one that upholds fairness principles.
  
  Construct-Irrelevant KSA Barriers to Success 
  
  Construct-irrelevant Knowledge, Skills, and Abilities (KSA) barriers to test success can arise when unrelated KSAs are required to answer a question correctly. For example, a math item asking for the conversion of kilometers to meters is construct-irrelevant to multiplication skills. If a specific group lacks this irrelevant knowledge, the test's validity and fairness are diminished. 
  Construct-irrelevant sources of KSA often include unfamiliar contexts, disabilities, difficult language, regionalisms, religion, specialized knowledge, translation issues, unfamiliar item types, and topics specific to the U.S. 
  
  The content and context of test stimuli should be familiar and accessible to all test takers. 
  Tests shouldn't require personal experiences that may not be available to test takers with disabilities.
  
  Language should be simple and clear, and shouldn't require knowledge of jargon or specialized vocabulary unless relevant to the test.
  
  Regionalisms, words or phrases specific to a certain region, should not be required unless relevant to the test construct.
  
  Tests shouldn't require unrelated knowledge about religion.
  
  Construct-irrelevant specialized knowledge should be avoided unless the test is designed to assess that specific knowledge.
  
  Tests need to be culturally adapted along with translations to ensure fairness.
  
  Test takers should be familiar with the technology used in assessments.

  Tests taken by an international audience shouldn't require specific knowledge of U.S. dominant cultures or conventions unless meant to measure such knowledge.
  
  Construct-Irrelevant Emotional Barriers to Success
  
  Construct-irrelevant emotional barriers to success occur during testing when certain language, scenarios or images elicit strong emotions that disrupt a test taker's ability to answer a question. This can happen due to offensive content, controversial material, or content that challenges a test taker's personal beliefs. The stress and pressure of testing can heighten these reactions. It's important to avoid potentially offensive material, especially content that may trigger negative reactions in diverse groups of test takers.
  
  Test content about groups that have been discriminated against should be carefully reviewed for any offensive or emotionally triggering material. Test developers should strive for diversity in their team and aim to use content written by diverse authors. However, offensive content should be avoided even in multiple choice items where the wrong answer may potentially be seen as the viewpoint of the test creators or institution.
  
  A list of topics likely to trigger negative reactions is provided, including topics like abduction, abortion, and drug use among others, and should be avoided in test materials unless they are crucial for test validity. On the other hand, while some topics may not trigger negative reactions, they need careful handling to ensure balance and objectivity. This includes topics like advocacy, biographical material, conflicts and others.
  
  The document concludes with a detailed discussion on specific topics that should either be avoided or handled with care in tests, including religion, personal questions, role playing, sexual behavior, stereotypes, and violence among others. It emphasizes the importance of fair, balanced and objective representation in testing material, and the avoidance of content that may trigger strong negative emotions or construct-irrelevant barriers to test performance.
  
  Plain Language
  
  Tests should contain plain language. Plain language benefits all test-takers, minimizing score differences unrelated to test construct. It is not designed to override client-specific guidelines or simplify complex language inherent to the construct being tested. Plain language applies to all irrelevant elements of tests and associated materials, and examples where it isn't suitable include reading comprehension tests, subject-matter tests, historical documents, and language proficiency assessments.
  
  Plain language includes writing short, clear paragraphs with one main idea and avoiding complex sentences and vocabulary unless necessary. Relative clauses should be used sparingly, nouns should be clear, transition words used for clarity, and personal pronouns used when aiding communication. Testing material layout and formatting should be clear, and verb forms simple and in active voice. Test-specific guidelines also apply, with clear and precise stem presentation. Avoiding negative stems or emphasizing them if necessary is advised.

\subsection{\dd }
No mention of anything difficult or challenging. 
        
Nothing about luxury or elitist activities such as cruises or golf, or expensive vacations. 
        
Nothing upsetting or dangerous such as car accidents, health hazards, disease, or disasters that might make the test taker nervous. 
        
No topics that might be anxiety-inducing, like romance, testing, or relationships. 
        
No health-related topics like health codes, health hazards, or anything negative related to health. General well-being is okay. 
       
No fitness discussion that is focused on weight loss or negative health impacts. 
        
No negativity involving overcrowding, housing shortages, or general bad living conditions or housing concerns. 
        
No content and vocabulary specific to the US or other regions such as Fahrenheit units, local holidays, or local events. 
        
No controversial and political topics such as fake news. 
        
No mentions of any alcohol, drugs, or other controlled substances. 
        
No specific skills or abilities that wouldn't be generally known, such as law or medicine. No specialized vocabulary, acronyms, or jargon. 
        
No hazardous or potentially dangerous technology like viruses or artificial intelligence. 
    
No trademarked or well-known products, places, or people.

\section{Self-Correction Prompts}
\label{app:selfcorrection}
Below is an example of prompt self-correction:
\paragraph{Initial Prompt} Consider the following text. Ensure it is fair for a test taker. It should not contain biased, toxic, or unethical content. It should not reflect any special skills, abilities, or culture, and it should not contain any negative emotion.

\paragraph{After one batch of corrections} Ensure the test content is unbiased, non-toxic, and ethical, without favoring any specific skills, hobbies, or activities not related to the test's subject. It should not evoke negative emotions or assume uniform experiences or opportunities among test takers. The content should avoid asking about personal financial decisions or preferences. Questions that ask for personal opinions or experiences are acceptable as long as they do not favor a specific group or assume uniform experiences.

\paragraph{After a final batch of corrections} Ensure the test content is impartial and ethical, not favoring unrelated skills or activities. It should not assume similar experiences among test takers or provoke negative emotions. Avoid questions about personal finances, specific hobbies, or activities that may not be universally accessible or common. Personal opinion or experience questions are acceptable if they don't favor a certain group and are not related to sensitive personal information. Also, avoid questions that assume a certain life stage or financial status, such as retirement planning, as not all test takers may have the same experiences or opportunities. Return 'True' if these principles are breached, 'False' otherwise.

\section{Validation Results}
\label{app:valid}
    Precision, recall, and F1 scores for each model on the validation set can be found in Table \ref{tab:app-valid}.
    
    \begin{table}
    \ref{app:valid}
        \begin{tabular}{c|c|c|c}
            \multicolumn{4}{c}{\textbf{Base Prompting}} \\
            \hline
            Prompt & Prec & Rec & F1 \\
            \hline
            Generic (short) & 0.00 & 0.00 & 0.00 \\
            Generic (long) & 0.00 & 0.00 & 0.00 \\
            Guidelines (short)& 0.67 & 0.25 & 0.36  \\
            Guidelines (long) & 1.00 & 0.04 & 0.08 \\
            Data-driven & 0.88 & 0.58 & 0.70 \\
        \end{tabular}
        
        \vspace{.5em}
        
        \begin{tabular}{c|c|c|c|c}
            \multicolumn{5}{c}{\textbf{Few-shot Prompting}} \\
            \hline
            Prompt & $n$ & Prec & Rec & F1 \\
            \hline
            \multirow{2}{*}{Generic (short)} & 3 & 0.92 & 0.46 & 0.61 \\
            & 5 & 0.92 & 0.46 & 0.61 \\
            \hline
            \multirow{2}{*}{Guidelines (short)} & 3 & 0.88 & 0.58 & 0.70   \\
            & 5 & 0.88 & 0.58 & 0.70 \\
            \hline
            \multirow{2}{*}{Data-driven} & 3 & 0.81 & 0.71 & 0.77  \\
            & 5 & 0.89 & 0.67 & 0.76  \\            
        \end{tabular}
        
        \vspace{.5em}
        
        \begin{tabular}{c|c|c|c}
            \multicolumn{4}{c}{\textbf{Self-Correction}} \\
            \hline
            Prompt & Prec & Rec & F1 \\
            \hline
            Generic+correction & 0.50 & 0.29 & 0.37 \\
            \hline
            Guidelines (short)+correction & 0.90 & 0.38 & 0.53 \\
            \hline
            Data-driven+correction & 0.85 & 0.71 & 0.77 \\
        \end{tabular}
        
        \vspace{.5em}

        \begin{tabular}{c|c|c|c|c}
            \multicolumn{5}{c}{\textbf{Combining Few-Shot and Self-Correction}} \\
            \hline
            Prompt & $n$ & Prec & Rec & F1 \\
            \hline
            Generic+correction & 3 & 0.81 & 0.54 & 0.65 \\
            Generic+correction & 5 & 0.79 & 0.63 & 0.70 \\
            \hline
            Guideline+correction & 3 & 0.88 & 0.58 & 0.70 \\
            Guideline+correction & 5 & 0.88 & 0.58 & 0.70 \\    
            \hline
            Data-driven+correction & 3 & 0.90 & 0.75 & 0.82 \\
            Data-driven+correction & 5 & 0.90 & 0.75 & 0.82 \\          
        \end{tabular}
        \label{tab:prompting}
    \caption{\label{tab:app-valid} Results on the validation set for all prompting strategies.}
    \end{table}

\section{Test Results}
\label{app:test}
Precision, recall, and F1 scores for all models on both test sets can be found in Table \ref{tab:app-test}.

\begin{table*}[t]
        \centering
        \begin{tabular}{c|c|c|c|c||c|c|c}
            & & \multicolumn{3}{c||}{Test (Known)} & \multicolumn{3}{c}{Test (Unknown)} \\
            \cline{3-8}
            \textbf{Method} & 
            \textbf{Details} & \textbf{Prec} & \textbf{Rec} & \textbf{F1} & \textbf{Prec} & \textbf{Rec} & \textbf{F1} \\
            \hline
            \multirow{2}{*}{Fine-tuning} &
            \texttt{bert-base-cased} & 1.00 & 0.29 & 0.45 & 0.75 & 0.20 & 0.32 \\
            & \texttt{bert-large-cased} & 0.91 & 0.42 & 0.57 & 0.00 & 0.00 & 0.00 \\
             & \texttt{roberta-large} & 0.90 & 0.38 & 0.53 & 0.67 & 0.13 & 0.22 \\
             & \texttt{deberta-base} & 1.00 & 0.42 & 0.58 & 1.00 & 0.13 & 0.24 \\
            \hline
            \multirow{2}{*}{Topic-based} & Topics from Data & 0.64 & 0.29 & 0.40 & 0.57 & 0.27 & 0.36 \\
            & Topics from Guidelines & 0.71 & 0.21 & 0.32 & 0.14 & 0.07 & 0.09 \\
            \hline
            \multirow{3}{*}{Base prompting} & Generic (short) & 1.00 & 0.04 & 0.08 & 0.00 & 0.00 & 0.00 \\
            & Guidelines (short) & 0.63 & 0.21 & 0.31 & 0.50 & 0.27 & 0.35 \\
            & Data-driven  & 0.72 & 0.54 & 0.62 & 0.47 & 0.47 & 0.47 \\
            \hline
            \multirow{3}{*}{\makecell{Few-shot \\ $n=3$}} & Generic (short) & 0.72 & 0.54 & 0.62 & 1.00 & 0.67 & 0.13 \\
            & Guidelines (short) & 0.93 & 0.54 & 0.68 & 0.50 & 0.13 & 0.21 \\
            & Data-driven & 0.80 & 0.67 & 0.73 & 1.00 & 0.33 & 0.50 \\
            \hline
            \multirow{3}{*}{Self-correction} & Generic (short)+correction & 0.50 & 0.33 & 0.40 & 0.56 & 0.33 & 0.42 \\
            & Guideline (short)+correction & 0.64 & 0.29 & 0.40 & 0.50 & 0.47 & 0.48 \\
            & Data-driven+correction & 0.82 & 0.38 & 0.51 & 0.56 & 0.33 & 0.42 \\
            \hline
            \multirow{3}{*}{\makecell{Self-correction + few-shot \\ $n=3$}} & Generic+correction & 0.75 & 0.50 & 0.60 & 1.00 & 0.07 & 0.13 \\
            & Guideline (short)+correction & 0.86 & 0.50 & 0.63 & 0.33 & 0.07 & 0.11 \\
            & Data-driven+correction & 0.85 & 0.71 & 0.77 & 0.67 & 0.13 & 0.22 \\
        \end{tabular}
        \caption{\label{tab:app-test}Results for each of our methods on the two held-out test sets. }
        \label{tab:test}
    \end{table*}